\documentclass[conference]{IEEEtran}
\IEEEoverridecommandlockouts

\usepackage{cite}
\usepackage{personal_import}
\usepackage{amsmath,amssymb,amsfonts}
\usepackage{algorithmic}
\usepackage{graphicx}
\usepackage{textcomp}
\usepackage{xcolor}
\usepackage{graphicx}
\usepackage{float}
\usepackage{caption}
\usepackage{url}
\def\BibTeX{{\rm B\kern-.05em{\sc i\kern-.025em b}\kern-.08em
    T\kern-.1667em\lower.7ex\hbox{E}\kern-.125emX}}
\begin{document}

\title{SurgBox: Agent-Driven Operating Room Sandbox with Surgery Copilot} 

\author{\IEEEauthorblockN{Jinlin Wu\textsuperscript{1,2$\dag$}, Xusheng Liang\textsuperscript{1$\dag$}, Xuexue Bai\textsuperscript{3}, Zhen Chen\textsuperscript{1*}}
\IEEEauthorblockA{
\textsuperscript{1}{CAIR, HKISI, CAS, Hong Kong SAR} \\
\textsuperscript{2}{MAIS, Institute of Automation, Chinese Academy of Sciences, China} \\
\textsuperscript{3}{Peking Union Medical College Hospital, China}
\thanks{$\dag$\,Equal contribution, $*$\,Corresponding author.}
\thanks{This work is supported by the InnoHK program, and the National Natural Science Foundation of China (Grant No.\#62306313).}
}

}

\maketitle

\begin{abstract}
Surgical interventions, particularly in neurology, represent complex and high-stakes scenarios that impose substantial cognitive burdens on surgical teams. Although deliberate education and practice can enhance cognitive capabilities, surgical training opportunities remain limited due to patient safety concerns. To address these cognitive challenges in surgical training and operation, we propose SurgBox, an agent-driven sandbox framework to systematically enhance the cognitive capabilities of surgeons in immersive surgical simulations. Specifically, our SurgBox leverages large language models (LLMs) with tailored Retrieval-Augmented Generation (RAG) to authentically replicate various surgical roles, enabling realistic training environments for deliberate practice. In particular, we devise Surgery Copilot, an AI-driven assistant to actively coordinate the surgical information stream and support clinical decision-making, thereby diminishing the cognitive workload of surgical teams during surgery. By incorporating a novel Long-Short Memory mechanism, our Surgery Copilot can effectively balance immediate procedural assistance with comprehensive surgical knowledge. Extensive experiments using real neurosurgical procedure records validate our SurgBox framework in both enhancing surgical cognitive capabilities and supporting clinical decision-making. By providing an integrated solution for training and operational support to address cognitive challenges, our SurgBox framework advances surgical education and practice, potentially transforming surgical outcomes and healthcare quality. The code is available at \href{https://github.com/franciszchen/SurgBox}{https://github.com/franciszchen/SurgBox}.

\end{abstract}


\begin{figure*}[h]
    \centering
    \includegraphics[width=0.92\linewidth]{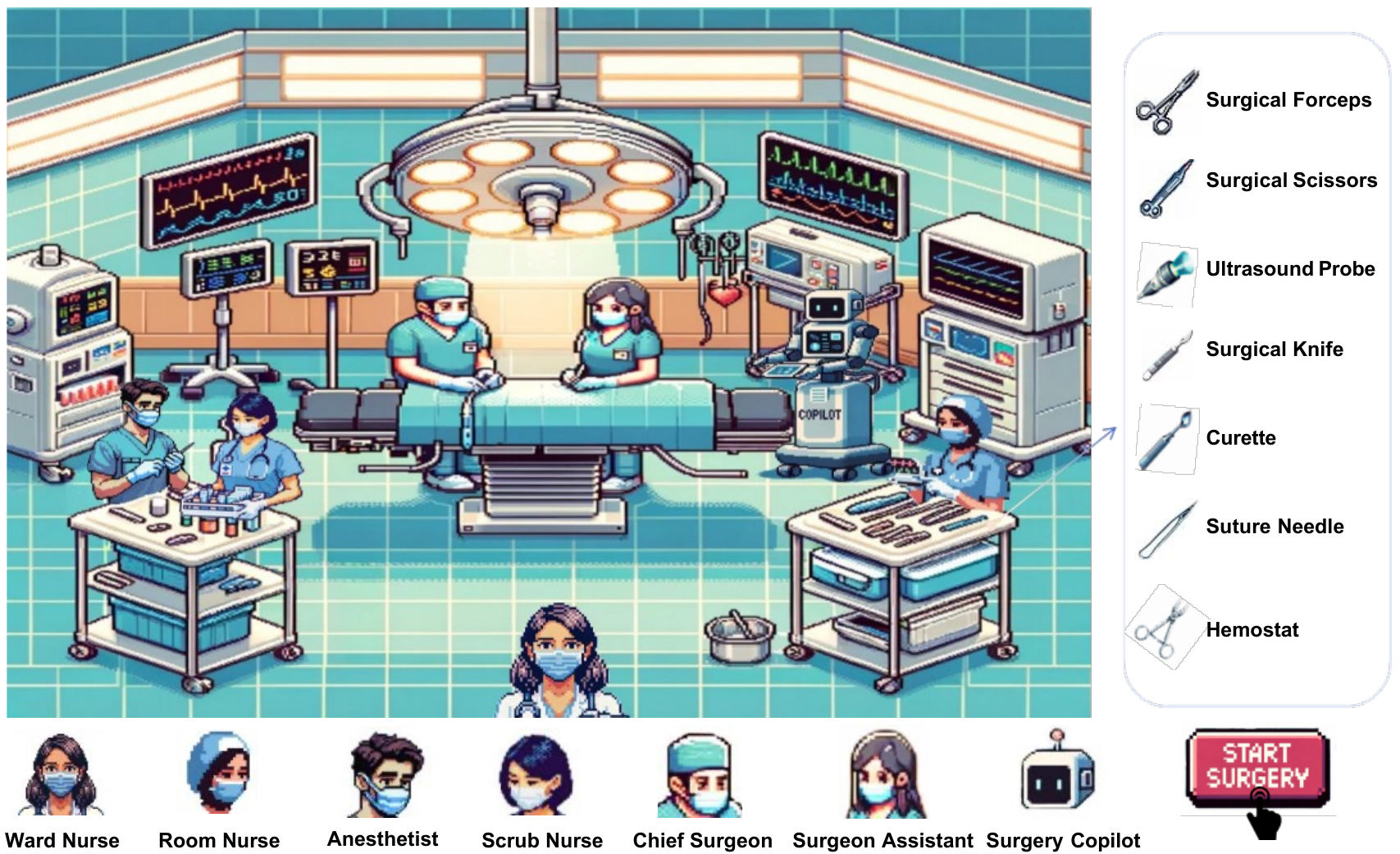}
    \caption{The overview of the SurgBox framework. In the simulated operating room sandbox,  each surgical role is automatically driven by an LLM-based agent. In particular, we devise the Surgical Copilot that is responsible for planning and guiding the entire surgical process.}
    \label{fig:Overview}
\end{figure*}

\section{Introduction}
Surgical interventions represent some of the most complex and high-stakes scenarios in medicine, with outcomes directly impacting treatment efficacy and life quality \cite{Varghese2024_nature}. In particular, neurosurgical interventions involve intricate workflows and critical decision-making throughout complicated surgical stages \cite{varghese2024artificial,chen2023surgical,luo2024surgplan,chen2023surgicalbibm}. This procedural complexity creates substantial cognitive burdens for surgical teams, who must simultaneously process multiple information streams while maintaining precision in their actions. Studies have shown that such cognitive demands significantly increase the risk of surgical errors \cite{frederiksen2020cognitive,chen2024asi}, potentially leading to devastating patient outcomes \cite{knight2021global}.

Although deliberate education practice has been proven effective in enhancing cognitive capabilities \cite{mao2021immersive}, safety and ethical concerns severely limit surgeons' opportunities to practice in actual surgical procedures \cite{aouicha2022patient}, especially for high-risk or rare conditions. Therefore, leveraging advanced AI techniques to coordinate surgical information stream and support clinical decision-making shows great promise in addressing cognitive challenges from both training and operational perspectives.

Recent AI advances have sparked a growing interest in developing safe, controllable virtual environments through generative agents to simulate real-world scenarios. Smallville \cite{park2023generative} introduces architectural and interaction patterns for enabling believable simulations of human behavior, and MetaGPT \cite{hongmetagpt} provides solutions for collaborative software engineering by imitating efficient human workflows. In the healthcare domain, pioneer studies (\textit{e.g.}, AI Hospital \cite{fan2024ai} and MedAgents \cite{tang2023medagents}) have explored using large language models (LLMs) agents to mimic clinical roles, interactions and decision-making. Concurrently, Agent Hospital \cite{li2024agent} aims to build AI systems that involve interactive pipelines across various medical scenarios.

Inspired by these LLM-based simulation methods, we introduce SurgBox, an innovative agent-driven framework designed to systematically enhance cognitive capabilities of surgeons through immersive surgical simulations. Our framework employs LLM agents enhanced with specialized Retrieval-Augmented Generation (RAG) banks to authentically replicate various surgical roles, including chief surgeon, assistant surgeon, nurses, and anesthetists. Through deliberate practice in these high-fidelity simulations, surgeons can develop automatic responses to various surgical scenarios, improving their ability to process complex information streams and make critical decisions under pressure. This repeated exposure to diverse surgical situations in a risk-free environment helps build robust cognitive schemas, enabling more efficient information processing and decision-making during actual procedures.

To amplify the training benefits of SurgBox and further reduce cognitive load during live surgeries, we devise the Surgery Copilot, the first AI-driven assistant designed to actively support surgical decision-making and workflow management in real time. This specialized agent helps surgeons maintain situational awareness by effectively coordinating and filtering information streams, providing contextually relevant guidance, and proactively identifying potential risks before they escalate into complications. By implementing a Long-Short Memory mechanism within the Surgery Copilot, we enable it to leverage knowledge from an extensive range of surgical cases while maintaining focus on immediately relevant surgical information. This balanced approach allows Surgery Copilot to deliver precisely targeted assistance throughout the procedure, significantly reducing the cognitive burden on surgical teams and potentially improving patient outcomes.

To validate the efficacy of our integrated approach, we conduct extensive experiments leveraging real neurosurgical procedure records of $128$ patients, encompassing a diverse range of conditions including various types of pituitary adenomas. Notably, our SurgBox reveals superior performance, achieving accuracy rates of 88.00\% and 88.02\% in surgical route selection and planning tasks respectively, outperforming state-of-the-art LLMs. Our system maintains high performance across all surgical stages, exhibits particular strength in diagnosing specific conditions, and performs well even with smaller sample sizes, showcasing its robustness in handling increasingly complex scenarios. These findings underscore the capability of our approach to effectively enhance surgical cognitive performance through both preparatory training and real-time assistance. By creating high-fidelity simulations for cognitive skill development while providing intelligent operation support, our system opens new frontiers for improving surgical outcomes and overall healthcare quality.



The contributions of this work are summarized as follows:
\begin{itemize}
    \item We propose SurgBox, the first agent-driven sandbox framework that systematically enhances surgeons' cognitive capabilities through immersive surgical simulations and deliberate practice.

    \item We develop a novel role-playing strategy using LLM agents enhanced with tailored RAG knowledge, precisely modeling the behaviors and interactions of all surgical team members.

    \item We design the Surgery Copilot, an AI-driven assistant with a Long-Short Memory mechanism, to coordinate the surgical team, provide guidance, and optimize learning throughout simulated procedures.

    \item Extensive experiments using real neurosurgical procedure records confirm the effectiveness of SurgBox in enhancing surgical cognitive and clinical decision-making capabilities of surgical teams across various conditions.       
\end{itemize}

\section{Related Work}

\subsection{LLM-Based Multi-Agent Framework}\label{AA}
Recent studies \cite{zhong2024debug} indicate that LLMs have been extensively used to simulate real-world dynamics and demonstrated their capability to interact in a competitive, dynamic real-world simulation, especially in fields such as epidemiology, sociology, and economics. LLM-based agents play a central role to enhance the predictive and analytical capabilities of models by simulating human decision-making processes \cite{eigner2024determinants}.

Many current agent-related works tend to simulate human-like understanding of human activities to generate human-like instructions\cite{zhang2024interactive} and facilitate complex interactions and decision-making in a wide range of situations\cite{yao2024tree,shinn2024reflexion}. For instance, game simulation \cite{xu2023exploring} verifies the ability of large models to participate in complex communication games. Software companies \cite{hong2023metagpt} employ Standard Operating Procedures (SOPs) to coordinate multi-agent systems based on LLMs, thus implementing metaprogramming technologies. Social simulation \cite{gao2023s} represents a pioneering step in the field of social network simulation utilizing LLM-based agents.

While these applications have shown the potential of LLMs in simulating real-world dynamics, there are still some areas that have not been fully explored. In the domain of medical and health simulation, existing research usually focuses on simulating treatment tasks rather than simulating the complete closed loop of treating patients' diseases\cite{teymourian2022closing}. This study aims to address this research gap by leveraging LLMs to enhance medical decision-making processes and improve the accuracy and efficacy of diagnosis and treatment protocols.

\subsection{Role-Play Agents}\label{BB}
Currently, open-source general-purpose foundational models have demonstrated impressive capabilities, and more and more work has begun to study models for role-playing. These models can adapt to complex environments \cite{barua2024exploring}, recall key information from long-term memory, and demonstrate continuous learning capabilities. \cite{yan2023larp} introduces a language agent role-playing framework that employs a modular approach for memory processing, decision-making, and interactive learning within the environment.\cite{tu2023characterchat} enhances the agent's capabilities through fine-tuning on a role-specific corpus.\cite{chen2024roleinteract} presents a systematic evaluation framework for role-playing agents, spanning from individual to group-level assessments.\cite{salemi2023lamp} augments the agent's role-playing capabilities by integrating retrieval enhancement techniques.

In the medical domain, AI Hospital \cite{fan2024ai} investigates the application of LLMs as clinical diagnosticians for real-time interactive consultation scenarios. However, ensuring robust diagnostic capabilities remains a significant challenge. The MedAgent-Zero \cite{li2024agent} approach employs a simulation system based on a knowledge base and LLM to model disease occurrence and progression, enabling doctor agents to accumulate experience from both successful and failed cases. Nevertheless, its efficacy in complex medical scenarios remains limited.

Our research endeavors to address the gap in the application of intelligent agents within the domain of clinical surgery. We have developed an advanced clinical intelligent surgery simulation system capable of managing complex surgical scenarios and simulating risk outcomes influenced by multiple clinical factors. This system facilitates intelligent agents in learning from both successful and suboptimal surgical cases, synthesizing lessons from failures, and continuously enhancing their capabilities, thus improving their proficiency in surgical decision-making and execution. The development of this comprehensive clinical surgery simulation system \cite{cardoso2023exploring} aims to advance the practical application of LLM technology in the medical domain and provide clinicians with more reliable decision-making support.







\begin{figure}[t]
    \centering
    \includegraphics[width=1.0\linewidth]{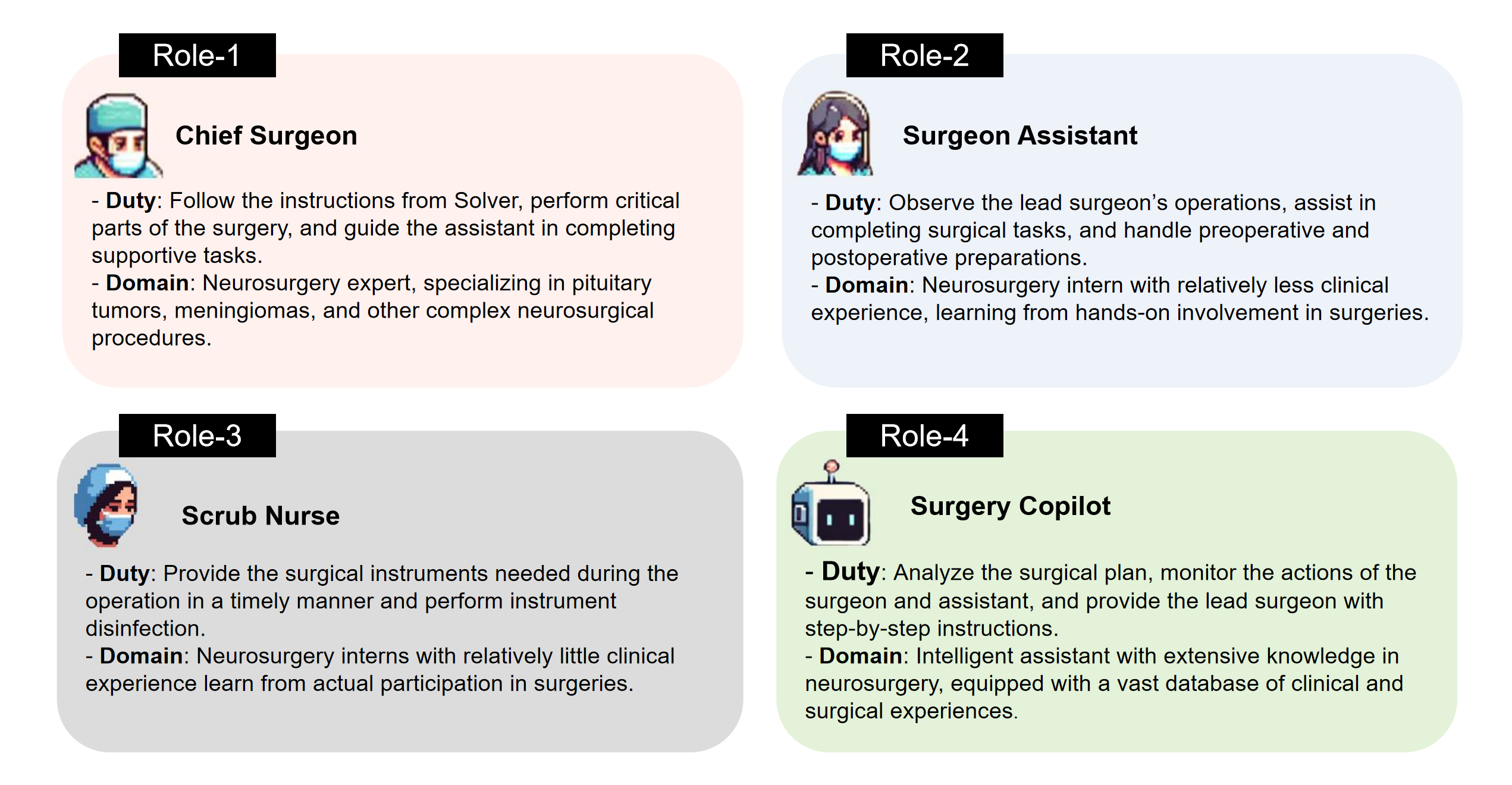}
    \caption{Examples of surgical roles in SurgBox include the chief surgeon, surgeon assistant, scrub nurse, and Surgery Copilot.}
    \label{fig:Example roles definition}
\end{figure}

\begin{figure*}[h]
    \centering
    \includegraphics[width=1.00\linewidth]{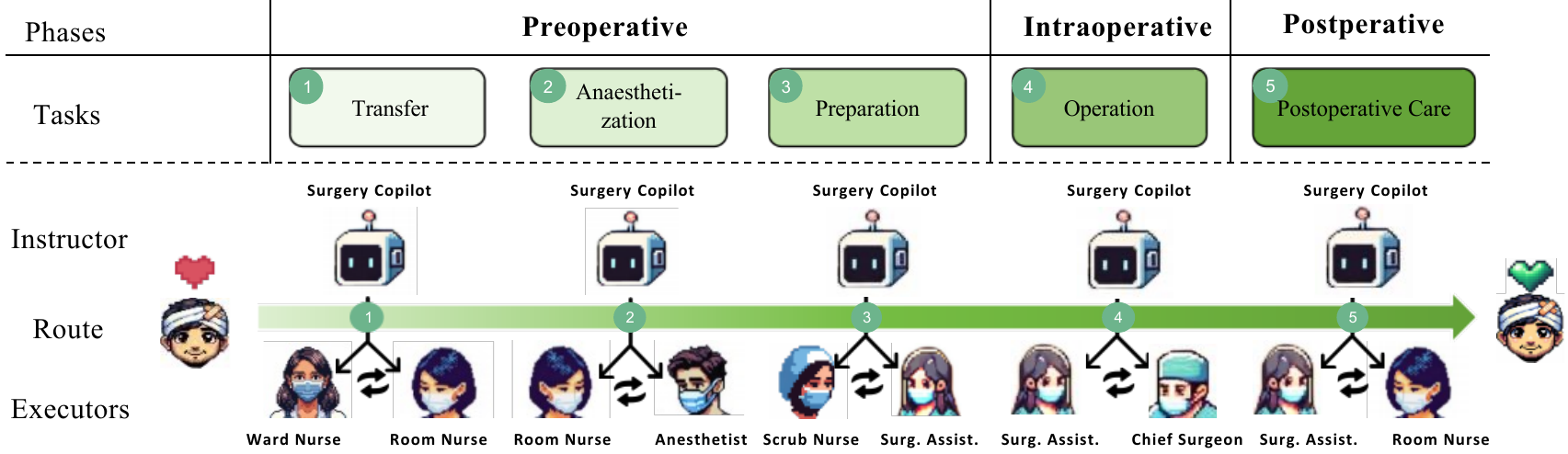}
    \caption{The surgical workflow in the SurgBox framework. The SurgBox framework simulates the patient's entire surgical closed loop, including patient transfer, anesthesia, preparation, surgical operation, and postoperative care.}
    \label{fig:Work Flow}
\end{figure*}

\section{Method}

In this section, we introduce SurgBox, a novel framework designed to comprehensively simulate the entire surgical process as shown in Fig. \ref{fig:Overview}. Sec. \ref{medical staff} elucidates the advanced role-playing methodology employed by SurgBox to accurately model the diverse medical professionals within an operating room environment. This approach meticulously captures the complex interactions and collaborative dynamics essential for completing surgical procedures. In Sec. \ref{copilot}, we present the Surgery Copilot, a pivotal component of SurgBox engineered to enhance surgical safety and optimize inter-professional collaboration efficiency. Sec.~\ref{memory} delineates the implementation of a sophisticated long-short memory mechanism integrated into the Surgery Copilot, which significantly improves the accuracy of surgical planning and interactions.

\subsection{Role Playing for SurgBox}
\label{medical staff}

\noindent \textbf{Role Playing of Medical Staffs.} As shown in Fig.~\ref{fig:Example roles definition}, SurgBox meticulously designs various surgical roles and employs LLM-based agents to accurately simulate the dynamics of operating rooms, establishing critical clinical roles including the patient, chief surgeon, surgeon assistant, scrub nurse, ward nurse, room nurse, and anesthetist. Each role's information and actions are generated using a tailored LLM, resulting in realistic and context-aware interactions. The characters take corresponding actions according to the stage task theme and event progression, engaging in dialogue with their counterparts. 

The SurgBox system employs a sophisticated role-playing mechanism that enables a realistic simulation of a surgical environment. This is achieved through a combination of predefined role-specific knowledge bases, contextual awareness, and dynamic interaction logic. Each role, such as Chief Surgeon or Anesthesiologist, is underpinned by a comprehensive database encompassing domain-specific terminology, procedures, and responsibilities. This allows the LLM to generate role-appropriate dialogues and behaviors. The system's contextual awareness tracks the progression of the surgery, ensuring that each role's responses align with the current surgical stage. Predefined interaction rules govern the communication patterns and hierarchical structures typical of a real operating room. The LLM's capacity for dynamic response generation, based on preceding dialogues and current circumstances, facilitates a natural flow of conversation. Integration of specialized medical and surgical vocabulary enhances the authenticity and professionalism of the generated dialogues. Each role is programmed to perform specific tasks and duties, such as the anesthetist monitoring vital signs or the nurse preparing instruments. The system also incorporates event-triggered mechanisms, where certain dialogues or actions may precipitate specific events, like changes in patient status, influencing subsequent responses from other roles. Furthermore, the LLM maintains conversational coherence by retaining and utilizing previous information in subsequent dialogues.

\noindent \textbf{Knowledge Enhancement of Roles.} To enhance the authenticity, effectiveness, and adaptability of the simulation, we implement an augmented domain knowledge retrieval and generation (RAG) method. This involves constructing a dedicated medical knowledge base for each key role in the operating room. The specialized knowledge bases encompass not only general medical knowledge but also focus on the specific expertise required by each role during the operation. For instance, the Chief Surgeon's knowledge base contains detailed surgical techniques, anatomical information, and complication treatment methods; the anesthetist focuses on anesthetic drug properties and patient monitoring protocols; and the nurse includes information on instrument preparation, aseptic techniques, and patient care. This refined knowledge base design ensures that each virtual role can access relevant and specialized knowledge support during the simulation, facilitating more accurate decision-making and actions.

\noindent \textbf{Interaction Between Roles.} As shown in Fig.~\ref{fig:Work Flow}, SurgBox structures each operation into distinct stages and subtasks to promote collaborative communication, facilitating multiple rounds of interaction between different roles and enabling the proposition and verification of solutions for each stage. The simulation encompasses the entire surgical process from preoperative to postoperative phases, segmented into three main stages: Preoperative, Intraoperative, and Postoperative. Each stage comprises specific tasks and participants, illustrating the patient's progression through key phases including transfer, anesthesia, surgical preparation, surgical operation, and postoperative care, with each phase involving the participation of corresponding medical personnel. This comprehensive approach not only replicates the surgical procedure itself but also incorporates preoperative preparation and postoperative care, providing a platform for medical teams to practice and optimize the entire surgical process in a virtual environment.

\noindent \textbf{Illustrative Example.} Consider a simulated neurosurgery procedure in SurgBox: During the preoperative phase, the Chief Surgeon reviews the patient's MRI scans and medical history, collaborating with the Anesthetist to develop a tailored anesthesia plan. The Scrub Nurse prepares the surgical instruments based on the procedure requirements. In the intraoperative phase, the Chief Surgeon performs the operation step by step according to the surgical plan, communicating with the Surgical Assistant for auxiliary operations. The anesthetist continuously monitors the patient's vital signs, adjusting anesthesia as needed, while the Scrub Nurse anticipates and provides necessary instruments. Post-surgery, the team transitions the patient to recovery, with Nurses monitoring vital signs and managing pain according to the anesthetist's instructions. Throughout this process, each role accesses its specialized knowledge base to inform decisions and actions, resulting in a highly realistic and educational simulation of the entire surgical experience.

\subsection{Surgery Copilot for SurgBox}
\label{copilot}

At the core of the SurgBox system lies the innovative Surgery Copilot, an advanced LLM-based assistant designed to orchestrate and optimize the entire surgical process. This Copilot serves as a central hub for coordination, planning, and support within the virtual operating room environment.

\begin{figure}[t]
    \centering
    \includegraphics[width=1.00\linewidth]{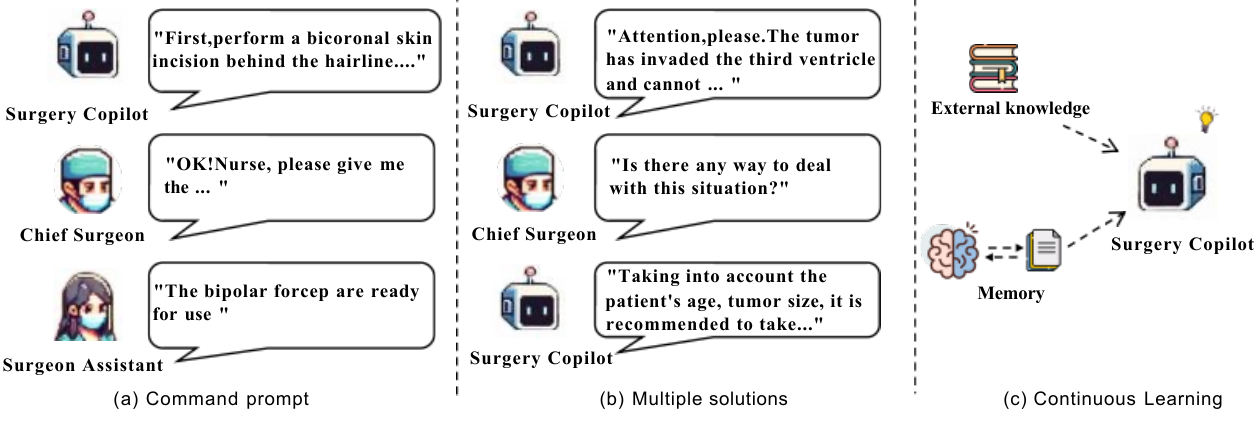}
    \caption{The interaction of Surgery Copilot and surgical roles in the surgical workflow.}
    \label{fig:Working mechanism of Surgery Copilot}
\end{figure}

\noindent \textbf{Role Playing.} The Surgery Copilot functions as an intelligent virtual assistant, seamlessly integrating into the SurgBox ecosystem to enhance surgical performance and outcomes. Its primary roles encompass real-time guidance, decision support, and adaptive learning as shown in Fig. \ref{fig:Working mechanism of Surgery Copilot}. The Copilot continuously monitors the surgical procedure, analyzing the operation of different surgery room roles to provide contextual insights and recommendations. It offers step-by-step guidance, alerts the team to potential risks, and suggests optimal techniques based on the current surgical context and patient-specific factors. The Copilot leverages a vast database of surgical experiences, constantly updated with the latest medical research and best practices, to offer evidence-based recommendations tailored to each unique surgical scenario. By integrating with other SurgBox components, the Copilot enhances team coordination, optimizes resource utilization, and provides real-time risk assessment. Its natural language interface allows seamless interaction with the surgical team, offering instant access to relevant information, from patient history to surgical plan. This comprehensive support system empowers surgeons to make informed decisions rapidly, potentially reducing procedural time and improving patient outcomes while serving as an invaluable educational tool for surgical trainees. 

\noindent \textbf{Interaction Between Copilot and Roles.} The Surgery Copilot orchestrates seamless interactions with various roles throughout the surgical process, ensuring efficient workflow management and optimal team coordination across three key phases. In the preoperative phase, it collaborates with surgeons to refine the surgical plan, assists anesthetists in reviewing the anesthesia strategy, and helps nurses prepare the operating room. During the intraoperative phase, the Copilot's interactions intensify, providing real-time guidance to surgeons, assisting scrub nurses with instrument preparation and aseptic techniques, supporting anesthetists in monitoring patient vitals, and coordinating with circulating nurses for equipment and supply management. In the postoperative phase, the Copilot aids in developing care plans, guides the monitoring of post-surgical complications, assists with pain management protocols, and facilitates a comprehensive debriefing session. Throughout all stages, the Copilot's continuous and tailored interactions with each role ensure clear communication, efficient coordination, and optimal patient care, significantly enhancing the overall efficacy of the surgical process by providing role-specific support and maintaining a cohesive team environment.

\noindent \textbf{Illustrative Example.} We consider a simulated neurosurgery procedure in SurgBox. The process unfolds as follows:
\begin{itemize}
    \item Preoperative: The Copilot analyzes the patient's MRI scans and medical history to generate a detailed surgical plan, briefs the team on potential challenges, and guides instrument preparation.
    \item Intraoperative: During the critical tumor resection, the Copilot provides real-time guidance to the surgeon, monitors vital signs, and quickly retrieves relevant case studies when an unexpected complication arises.
    \item Postoperative: The Copilot assists in formulating a detailed recovery plan, provides critical care instructions to the ICU team, and facilitates a debriefing session to highlight learning points.
    \item This comprehensive support demonstrates how the Surgery Copilot enhances team coordination, decision-making, and overall surgical outcomes within the SurgBox environment, providing an invaluable tool for medical training and simulation.
\end{itemize}

\subsection{Long-short Memory for Surgery Copilot}
\label{memory}

\begin{figure}[t]
    \centering
    \includegraphics[width=1.0\linewidth]{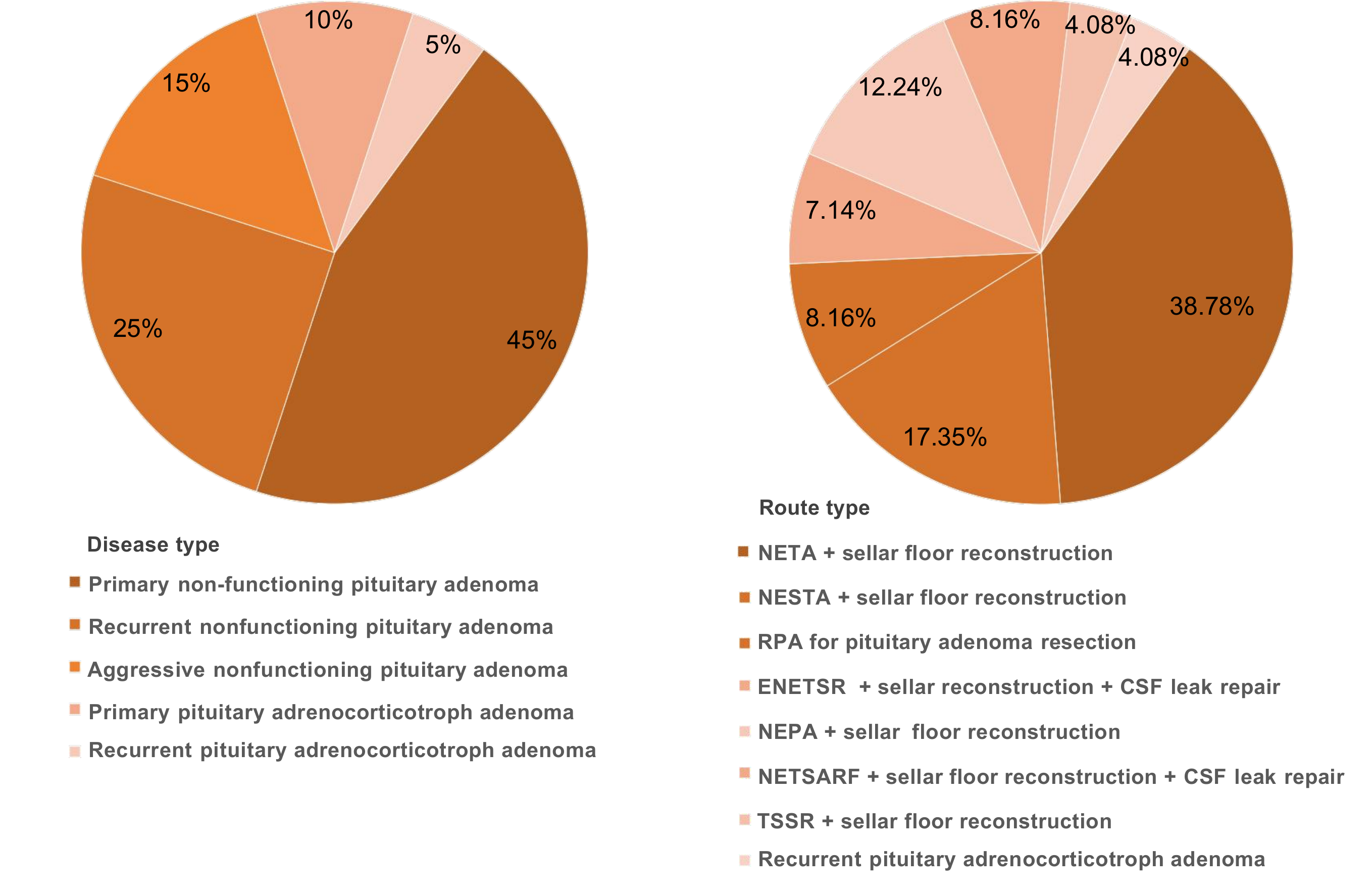}
    \caption{The distribution of neurosurgical diseases and neurosurgery routes in the real neurosurgical procedure records.}
    \label{fig:Data statistics}
\end{figure}

\begin{figure*}[t]
    \centering
    \includegraphics[width=0.8\linewidth]{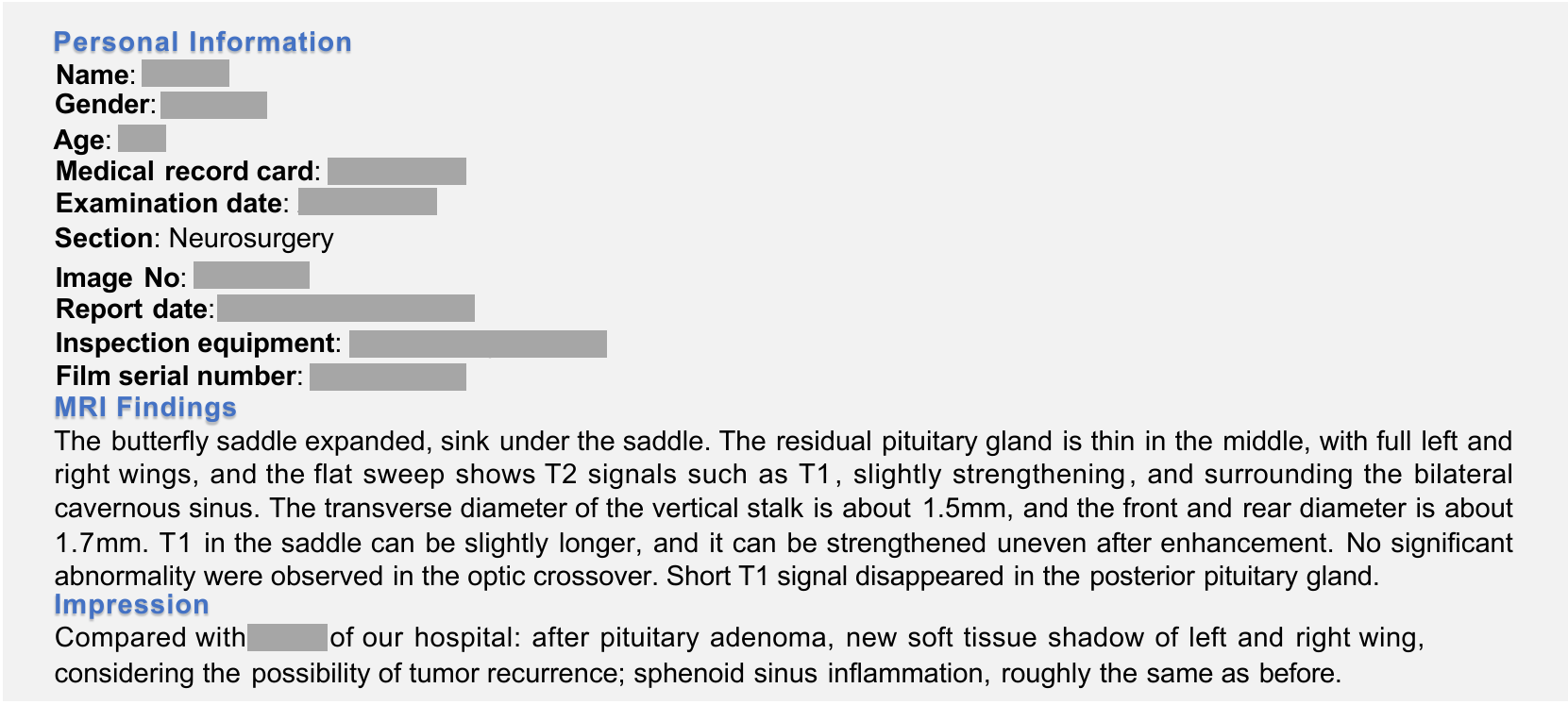}
    \caption{The example of MRI diagnostic reports.}
    \label{fig:MRI}
\end{figure*}

\begin{figure*}[t]
    \centering
    \includegraphics[width=0.9\linewidth]{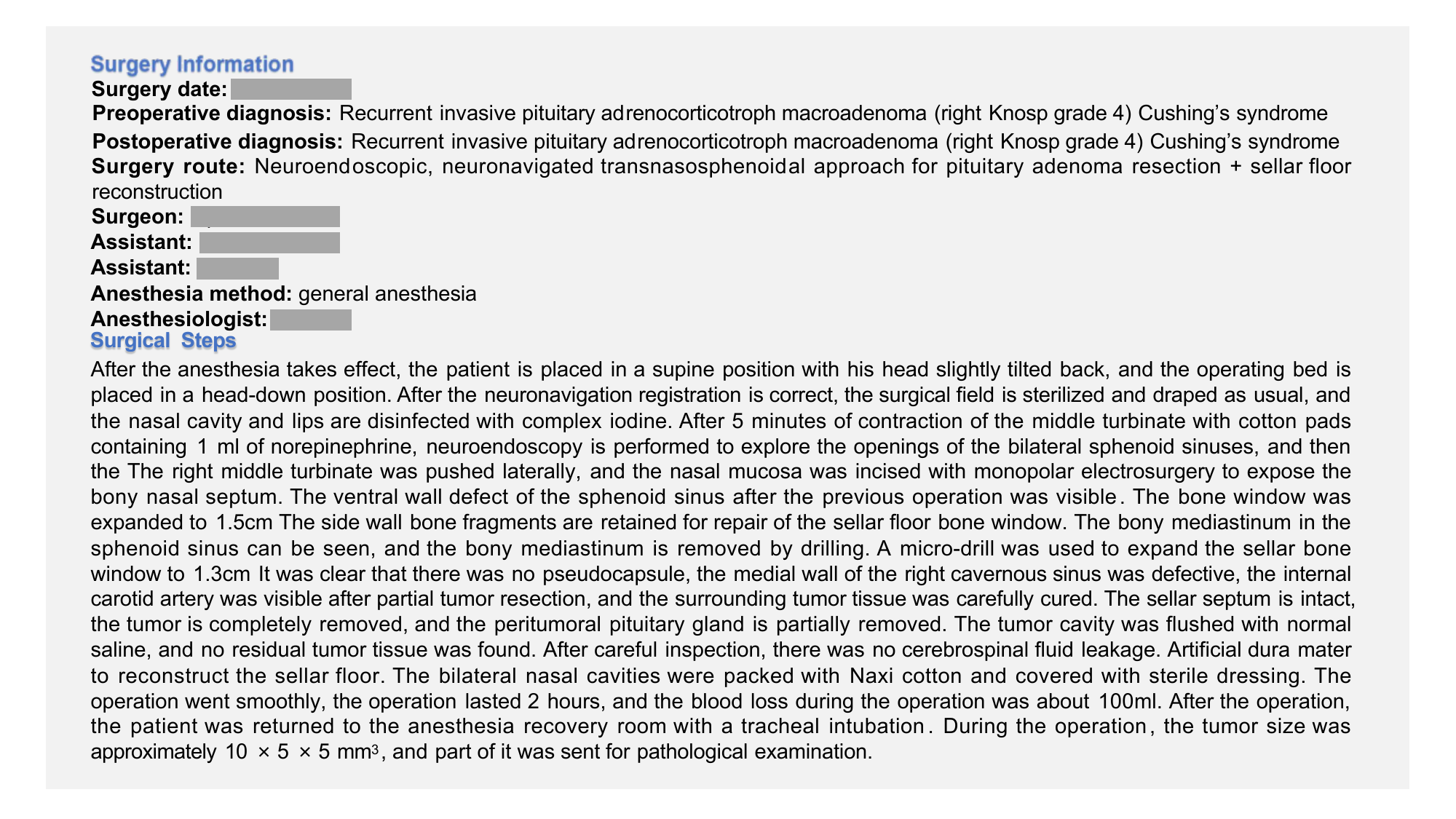}
    \caption{The example of neurosurgical procedure records.}
    \label{fig:Surgery Report}
\end{figure*}

\noindent \textbf{Short memory.} The Surgery Copilot is enhanced with a short-term memory mechanism that focuses on real-time information during the current surgery, including environmental data, inter-role dialogues, and ongoing operations. This feature enables the Copilot to swiftly retrieve relevant information about the corresponding stage and role during a single operation, ensuring timely and effective instructions without interference from historical data. The short-term memory significantly improves the Copilot's ability to respond rapidly to changes in surgical scenes and flexibly address emergencies. 


\noindent \textbf{Long memory.} Complementing the short-term memory, the long-term memory component of the Surgery Copilot continuously accumulates experience from multiple surgeries. This comprehensive knowledge base not only contains clear and complete records of each operation but also synthesizes experiential summaries, providing crucial guidance for future similar procedures. The long-term memory is continuously updated through an iterative learning process, involving thorough evaluation of each simulation, extraction of key lessons, and systematic integration of new insights. This mechanism allows the Copilot to build a rich repository of surgical knowledge and best practices over time.


We integrate short-term and long-term memory mechanisms to enhance Surgery Copilot's ability to iteratively improve its surgical planning and execution. While short-term memory is cleared after each operation to maintain focus, long-term memory is updated with new insights, allowing the Copilot to refine its strategies based on accumulated experience. This approach enables more comprehensive pre-operative planning by leveraging past procedures while maintaining agility in real-time support. The Copilot can analyze performance indicators, manage unexpected events, and identify best practices across various scenarios. As a result, Surgery Copilot becomes increasingly proficient at optimizing surgical workflows, anticipating complications, and providing guidance for each case.

\section{Dataset}



\subsection{Real Surgery Report Dataset} 
We collect the dataset of 128 real clinical surgery reports, as shown in Fig.~\ref{fig:Surgery Report}. To enhance physician agent performance, we have enriched the operative reports with additional contextual information, including basic information, patient history, and MRI findings in Fig.~\ref{fig:Data statistics}. This comprehensive dataset enables physician agents to practice and refine their decision-making skills. The dataset facilitates continuous improvement of agents' capabilities in complex environments that simulate real-world surgical scenarios. The ultimate goal is to equip physician agents with reliable, practical knowledge applicable to actual clinical situations.

\subsection{Simulated Surgery Report Dataset}
This dataset consists of 1,000 simulated surgical reports generated through multiple SurgBox simulation processes. It is based on different types of surgical procedures and comprehensive patient examination results, covering a wide range of information such as preoperative examination, surgical details, postoperative observations, and results. 


\section{Experiments}\label{sec:results}
In this section, we present the experimental details to validate the effectiveness of the proposed SurgBox framework and Surgery Copilot from various perspectives.

\subsection{Experimental Settings}
\noindent \textbf{Dataset.} We utilize real neurosurgery records as experimental datasets. These records contain detailed MRI analysis reports, surgical procedures, and clinical decision information. To protect patient privacy, all data were anonymized prior to analysis.
The dataset encompasses neurosurgeries employing various surgical methods. We divide the dataset into training and test sets. The training set is used for the initialization and optimization of the SurgBox system, while the test set is used to evaluate the performance of the system. 

\noindent \textbf{Evaluation.} To evaluate the performance of SurgBox, we employ two evaluation metrics specifically designed for SurgBox. 1) \textbf{Surgical Route Accuracy}: Evaluate the ability of LLM-based agents to select the best surgical route based on a given patient's condition. We compare the system's choices with the judgment of experienced neurosurgeons. 2) \textbf{Surgical Plan Accuracy}: Evaluate the ability of LLM-based agents to correctly plan and complete the entire surgery.

\noindent \textbf{Implementation Details.} The SurgBox system is built on advanced LLMs, and all our simulation experiments are based on the GPT-3.5-turbo-16k API. Specific prompt engineering techniques are employed to optimize the model's comprehension of medical terminology and surgical procedures. To simulate diverse surgical roles, dedicated knowledge bases, and behavior models are created for each role. As a core component, Surgery Copilot is allocated enhanced decision weights and expanded knowledge access. An iterative optimization method is adopted for the training process. The system initially underwent training on a limited number of surgical records, followed by a gradual expansion of the training set size, with continuous adjustment of model parameters and decision logic based on expert feedback. To ensure the repeatability and fairness of the experiments, we conducted all experiments in a fixed hardware environment and initialized the model with the same random seed.



\begin{table}[t]
	\centering

	\begin{tabular}{lcc}
		\toprule
		Model  &Surgery Route &Surgery Plan  \\
		\hline
                InternLM2\cite{cai2024internlm2} & 59.00 & 79.42  \\
                LLaMA 3\cite{meta2024introducing} & 64.00 & 80.33   \\
                GPT-3.5 & 72.00 & 82.26  \\
                GPT-4\cite{achiam2023gpt} & 79.00 & 85.68    \\
                Surgery Copilot & \textbf{88.00} & \textbf{88.02}   \\
                
		\hline
	\end{tabular}
 	\caption{Comparison of Surgery Copilot with diverse LLMs.}
	\label{tab:Performance of different model}
\end{table}

\subsection{Experimental Results and Analysis}

As shown in TABLE \ref{tab:Performance of different model}, our Surgery Copilot demonstrates superior performance in both Surgery Route and Surgery Plan categories, achieving scores of 88.00\% and 88.02\%, respectively. These results indicate that Surgery Copilot possesses substantial advantages in surgical route planning and surgical plan formulation.

The implementation of specific domain RAG technology substantially enhanced the performance of the baseline model, particularly in the Surgery Route category. This observation suggests that the incorporation of external knowledge retrieval can significantly enhance model performance in specialized domains. SurgBox demonstrated a distinct advantage over all other models by integrating a surgical field-specific knowledge base with the React method, effectively mitigating hallucinations and enhancing overall accuracy.

\begin{table}[b]
	\centering
	\scriptsize
        \setlength{\tabcolsep}{3pt}
	\begin{tabular}{lccccccccc}
		\toprule
		\multirow{2.5}{*}{\textbf{Model}}  &  \multicolumn{2}{c}{\textbf{Stage-1(25\%)}}  &  \multicolumn{2}{c}{\textbf{Stage-2(50\%)}} &  \multicolumn{2}{c}{\textbf{Stage-3(75\%)}} & \multicolumn{2}{c}{\textbf{Stage-4(100\%)}}\\
        \cmidrule(lr){2-3} \cmidrule(lr){4-5} \cmidrule(lr){6-7} \cmidrule(lr){8-9}
         & Comp & Acc  & Comp & Accuracy & Comp & Acc & Comp & Acc \\
        \midrule
                InternLM2 & 96.00 & 72.32 & 84.00 & 64.26 & 60.00 & 56.28 & 52.00 & 54.33  \\
                LLaMA 3 & \textbf{100.00} & 84.33 & 92.00 & 79.42 & 64.00 & 69.67 & 60.00 & 64.46  \\
                GPT-3.5 & \textbf{100.00} & 87.26 & 96.00 & 81.06 & 76.00 & 71.43 & 68.00 & 39.73 \\
                GPT-4 & \textbf{100.00} & \textbf{90.68} & 100.00 & \textbf{84.52} & 80.00 & 77.89 & 76.00 & 72.38  \\
                Surgery Copilot & \textbf{100.00} & \textbf88.00 & \textbf{100.00} & 83.98 & \textbf{92.00} & \textbf{80.06} & \textbf{84.00} & \textbf{77.19} \\
		\hline
	\end{tabular}
 	\caption{ Evaluation of completeness (Comp) and accuracy (Acc) on each model across different stages.}
	\label{tab:Comp and Acc}
\end{table}

The SurgBox shows remarkable performance in all stages from TABLE \ref{tab:Comp and Acc}. The system consistently maintained a superior completion rate, particularly in Stage 2 and 3. Concurrently, its accuracy remained elevated throughout all stages, with notably high and stable performance in the later stage, indicative of its robustness and reliability in complex surgical scenarios. Generally, the completion rate and accuracy of all models declined as stages progressed, reflecting the escalating complexity and challenges in the later phases of the surgical process. In contrast, while SurgBox exhibits a marginal decrease in completion rate, its accuracy diminished less significantly, demonstrating superior consistency and adaptability. This discrepancy underscores the advantages of models optimized for surgical scenarios, over general LLMs, particularly in managing complex late-stage surgical procedures. This trend emphasizes the significance and challenges associated with developing surgical simulation systems capable of maintaining high performance throughout the entire surgical process.


\begin{table}[t]
	\centering

	\begin{tabular}{lcc}
		\toprule
		Model  &Surgery Route &Surgery Plan  \\
		\hline
                Baseline & 72.00 & 83.69   \\
                \phantom{text}\textit{w/} Domain-RAG & 84.00 & 87.02   \\
                \phantom{text}\textit{w/} React\cite{yao2022react} & 72.00 & 80.56   \\
                \phantom{text}\textit{w/} Copilot\cite{wei2022chain} & 79.00 & 77.32   \\
                Surgery Copilot & \textbf{88.00} & \textbf{88.02}   \\
                
		\hline
	\end{tabular}
 	\caption{Ablation study of Surgery Copilot.}
	\label{tab:Ablation studys}
\end{table}

\begin{figure}[b]
    \centering
    \includegraphics[width=0.8\linewidth]{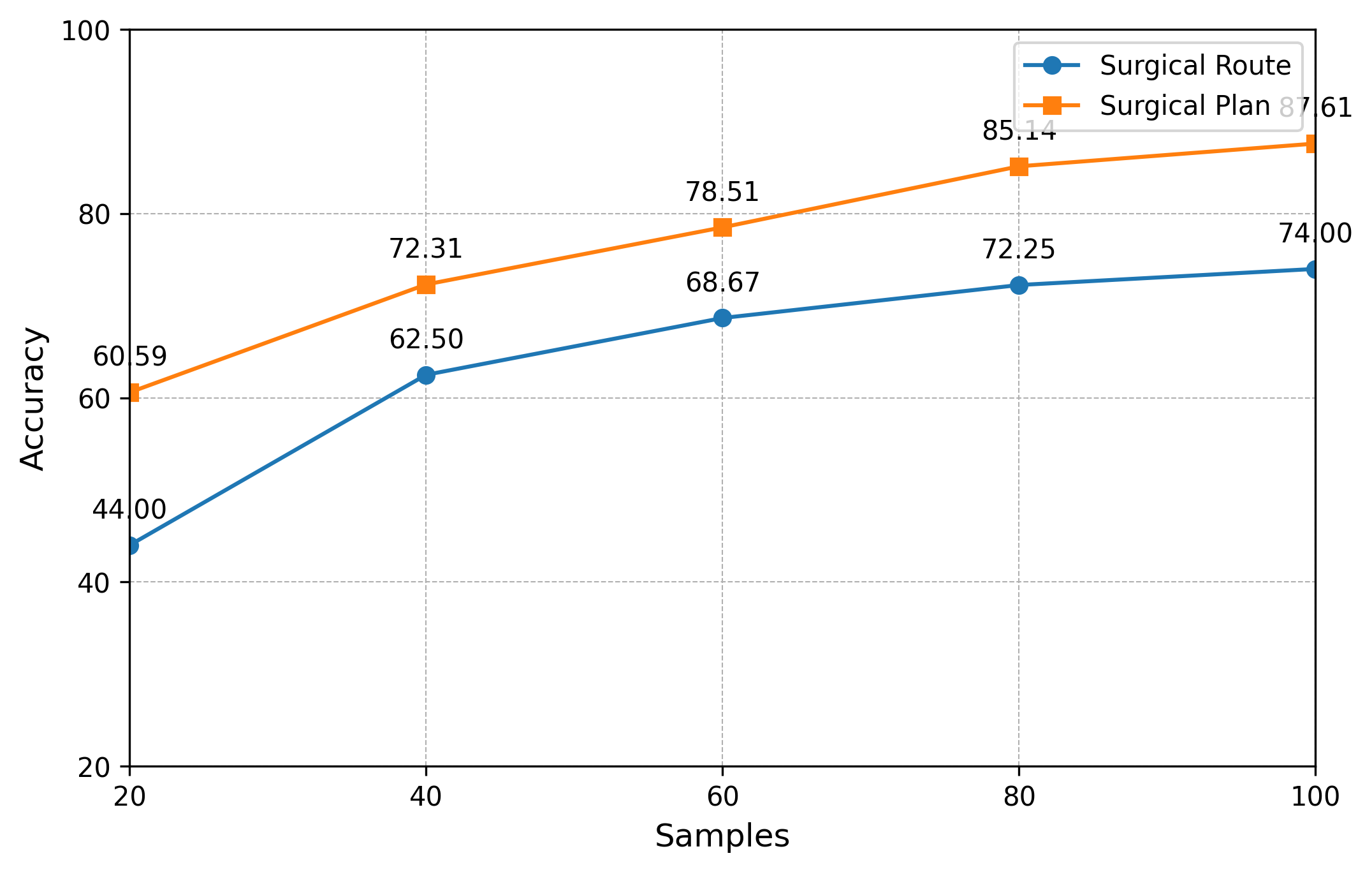}
    \caption{The performance of Surgery Copilot with different numbers of samples.}
    \label{fig:performance on the two tasks}
\end{figure}

We can summarize the following key conclusions from Fig.~\ref{fig:performance on the two tasks}. As the sample size increases, Surgery Copilot's performance in both surgical path and surgical planning tasks demonstrates significant improvement, exhibiting an upward but gradually plateauing trend. Notably, the model exhibits robust performance even with smaller sample sizes, which has substantial implications for real-world medical scenarios with resource constraints.

\begin{figure}[t]
    \centering
    \includegraphics[width=1.0\linewidth]{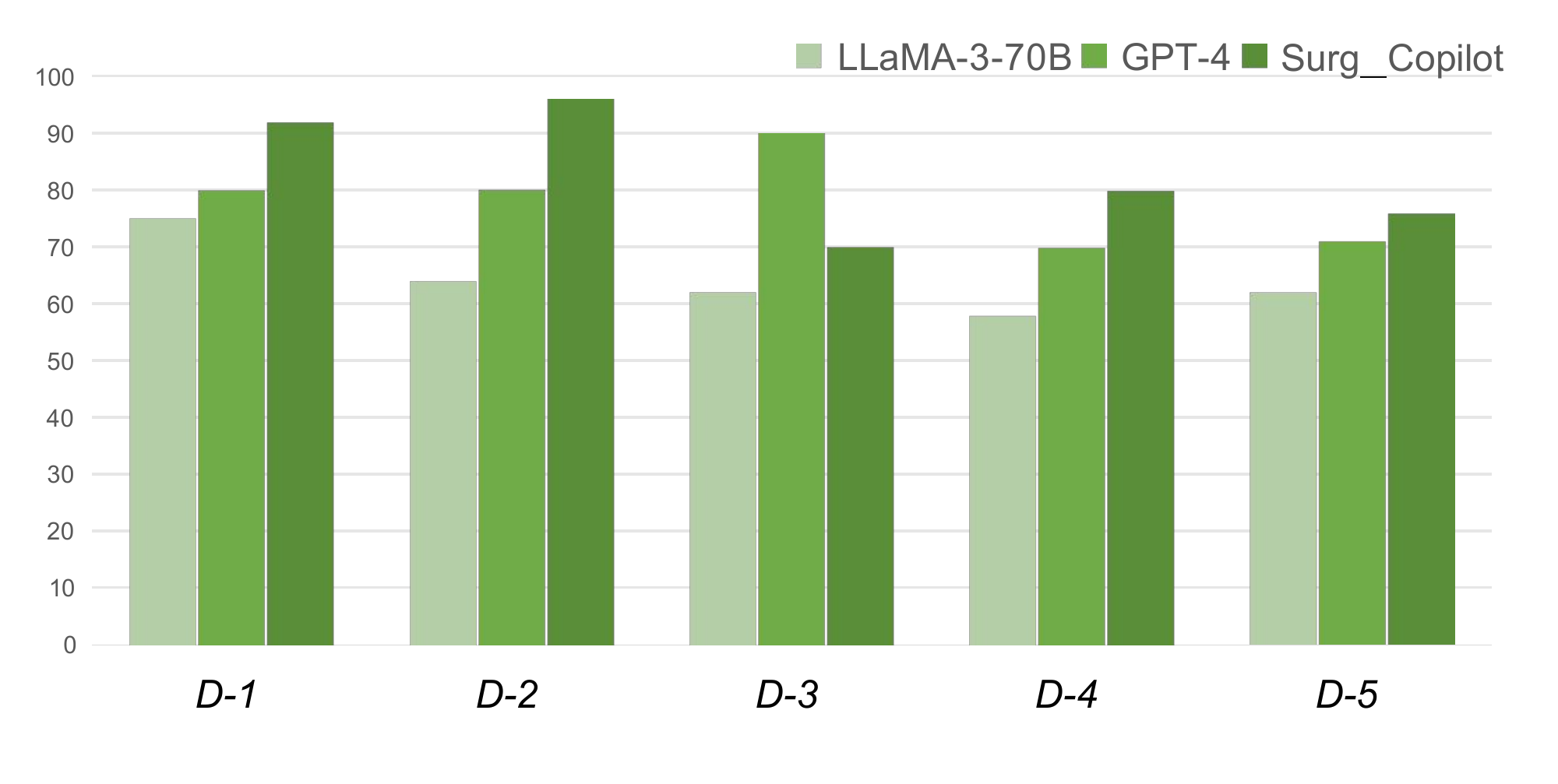}
    \caption{Comparison in specific diseases. D-1: Primary non-functioning pituitary adenoma,     
D-2: Recurrent nonfunctioning pituitary adenoma,
D-3: Aggressive nonfunctioning pituitary adenoma,    
D-4: Primary pituitary adrenocorticotroph adenoma,
D-5: Recurrent pituitary adrenocorticotroph adenoma,
}
    \label{fig:Accuracy of models in  distinct diseases.}
\end{figure}

Surgery Copilot performed well on primary non-functioning pituitary adenomas and recurrent pituitary corticosteroid adenomas, as illustrated in Fig.~\ref{fig:Accuracy of models in  distinct diseases.}, highlighting its advantage in diagnosing these specific diseases. The varied performance across different diseases suggests that the complexity and characteristics of each condition significantly influence the model's efficacy.

\begin{table}[t]
    \vspace{2mm}
    \centering
    \begin{tabular}{lc}
        \toprule
        Error Type & \# Num. \\
        \midrule
        Unable to cope with multiple intraoperative situations & 15 \\
        Misjudgment of the initial surgical approach & 24 \\
        Hallucinations about rare diseases & 5 \\
        \bottomrule
    \end{tabular}
        \caption{Mistake analysis for the failure during surgery.}
    \label{table:error_types}
\end{table}

The implementation of various technologies demonstrates differential impacts on the baseline model's performance in surgical planning and routing in Table \ref{tab:Ablation studys}. Notably, our specific domain RAG technology showed significant improvement, particularly in the Surgery Route category, increasing the score from 72.00\% to 84.00\%. This substantial enhancement of SurgBox suggests that integrating external knowledge retrieval and long-short term memory is highly effective in specialized domains, including surgery.

The Surgbox, which integrates a surgery-specific knowledge base with the React method, demonstrated superior performance across both categories, achieving 88.00\% for the surgery route accuracy and 88.02\% for the surgery plan accuracy. This demonstrates that the combination of domain-specific knowledge and advanced reasoning techniques effectively mitigates hallucinations and enhances overall accuracy in surgical planning and routing tasks.



\begin{figure*}[h!]
    \centering
    \includegraphics[width=0.87\linewidth]{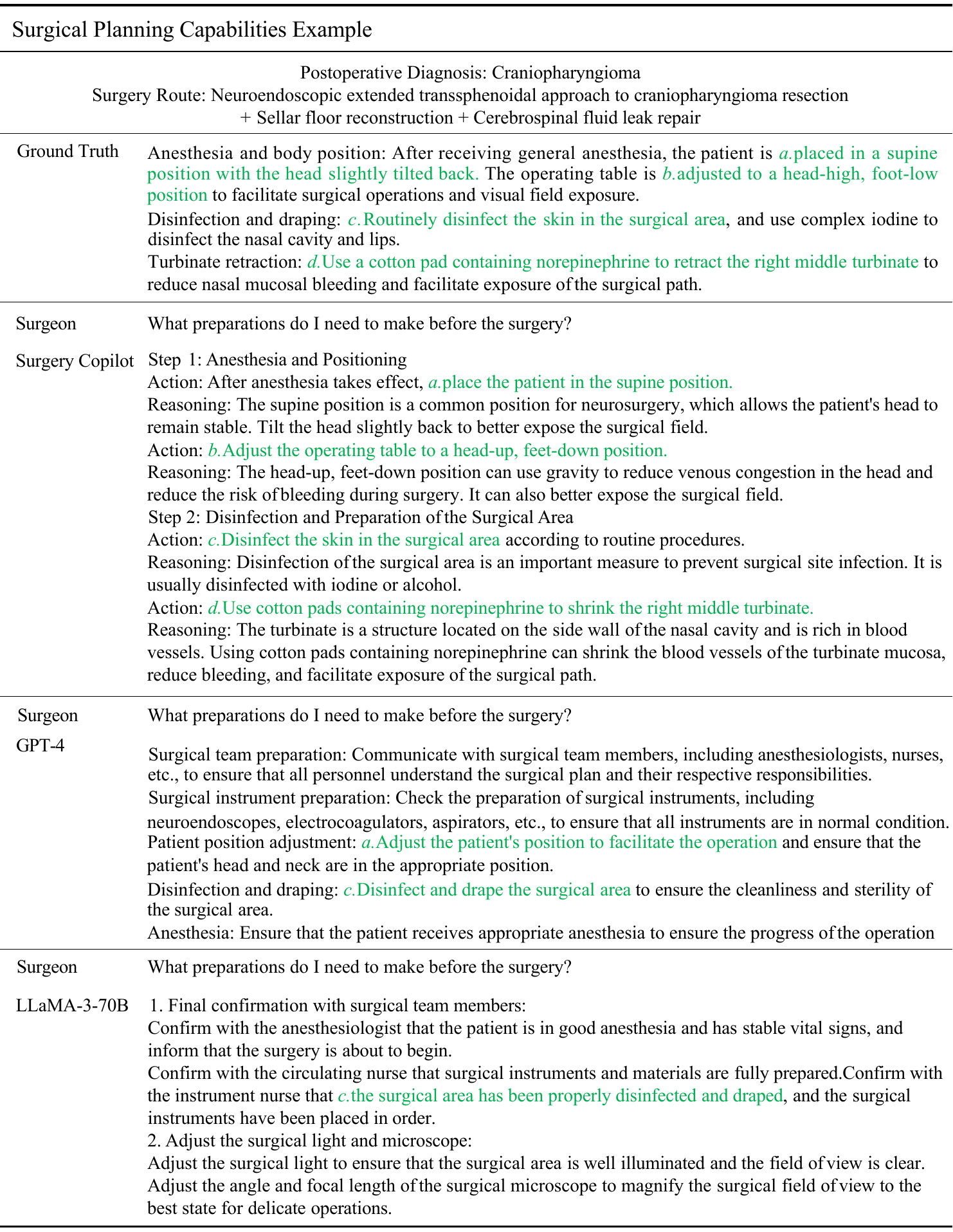}
    \caption{Example comparison of surgical video question-answering. We provided the answers generated by the  Llama-3-70B and GPT-4 for reference. The answers are based solely on extracted question information and surgery information. Surgery Copilot has a higher hit rate on key facts than others.}
    \label{fig:question-answering}
\end{figure*}



Through a detailed examination of the example presented in Fig.~\ref{fig:question-answering}, we can elucidate the distinct advantages of the Surgery Copilot. The surgery copilot demonstrates robust surgical planning capabilities, coupled with an impressive ability to decompose complex procedures into detailed, sequential steps, while exhibiting a profound understanding of specialized medical knowledge. Unlike GPT-4 and LLaMA-3-70B, the Surgery Copilot's responses are highly structured and targeted, meticulously dividing preparatory work into specific steps with clear action instructions and corresponding rationales. This structured response not only facilitates the surgical team's orderly completion of preparation tasks while minimizing errors but also serves as a valuable educational resource, enhancing comprehension of each step's importance. 



Despite its potential, Surgery Copilot exhibits limitations, as detailed in Table \ref{table:error_types}. A primary concern is the misclassification of the initial surgical approach, reflecting deficiencies in accurately assessing the extent of tumor invasion, patient-specific physiological and anatomical factors, and associated surgical risks. Furthermore, the system demonstrates the difficulty in managing concurrent intraoperative events, highlighting limitations in prioritizing and orchestrating appropriate responses to complex scenarios, exemplified by simultaneous cerebrospinal fluid leaks and hemorrhages. Additionally, the observed phenomenon of hallucinatory diagnoses pertaining to rare pathologies, wherein common intraoperative findings are misclassified, underscores limitations in the system's capacity for nuanced interpretation of subtle anatomical details and real-time intraoperative observations.

\section{Conclusion}
In this work, we present an integrated approach to address the critical cognitive challenges in surgical interventions. First, we propose SurgBox, an agent-driven sandbox framework that enables surgeons to systematically enhance their cognitive capabilities through deliberate practice in risk-free virtual environments. By leveraging LLM-based agents with tailored RAG knowledge banks, SurgBox creates highly realistic simulations of operating room dynamics, allowing surgeons to develop robust cognitive schemas for handling complex surgical scenarios. Second, we devise Surgery Copilot with the Long-Short Memory mechanism, to actively reduce cognitive load during live surgeries by providing intelligent information coordination and decision support. Extensive experiments validate the superiority of our approach in both surgical training and operational assistance. By enhancing cognitive capabilities while reducing cognitive burden, our work represents a significant advancement in surgical education and practice, potentially transforming surgical outcomes and healthcare quality.

{
\bibliographystyle{IEEEtran}
\bibliography{references}
}

\end{document}